\documentclass[11pt]{article}
\usepackage[margin=1in]{geometry}
\usepackage{graphicx}
\usepackage{multicol,color}
\usepackage{multirow}
\usepackage{arydshln}
\usepackage{amssymb}
\usepackage{mathtools}
\usepackage{enumitem}
\usepackage[square]{natbib}
\usepackage{authblk}
\usepackage{xcolor}
\usepackage{caption}
\definecolor{red}{HTML}{F44336}
\definecolor{green}{HTML}{4CAF50}
\definecolor{yellow}{HTML}{FFEE58}
\definecolor{blue}{HTML}{0D47A1}
\usepackage[colorlinks,linkcolor=blue,citecolor=blue]{hyperref}

\begin{document}
\title{Unified Representation of Genomic and Biomedical Concepts through Multi-Task, Multi-Source Contrastive Learning}

\author[1,2]{Hongyi Yuan\footnote{These authors contributed equally to this work.}}
\author[1]{Suqi Liu$^*$}
\author[3,4,5]{Kelly Cho}
\author[3,4,5]{Katherine Liao}
\author[3,4,5]{Alexandre Pereira}
\author[1,3]{Tianxi Cai}
\affil[1]{Department of Biomedical Informatics, Harvard Medical School}
\affil[2]{Department of Statistics and Data Science, Tsinghua University}
\affil[3]{VA Boston Healthcare System}
\affil[4]{Department of Medicine, Brigham and Women's Hospital}
\affil[5]{Department of Medicine, Harvard Medical School}
\date{}

\maketitle

\begin{abstract}
Biomedical concepts, along with genomic features such as single-nucleotide polymorphisms (SNPs) and gene expression, are crucial for understanding the genetic and biomedical relationships in modern medicine. Genome-wide association studies (GWAS), expression quantitative trait loci (eQTL), and phenome-wide association studies (PheWAS) derived from biobank data provide valuable insights into the interplay between genomic and clinical features. However, harmonizing existing biological and clinical knowledge from different sources to enable a better understanding of diseases and treatments remains challenging. Variations in biomedical concept coding systems and differences in SNP selections across data sources create barriers to effective integration.
To address this challenge, we introduce GENomic Encoding REpresentation with Language Model (GENEREL), a framework designed to bridge genetic and biomedical knowledge bases. What sets GENEREL apart is its ability to fine-tune language models to infuse biological knowledge behind clinical concepts such as diseases and medications. This fine-tuning enables the model to capture complex biomedical relationships more effectively, enriching the understanding of how genomic data connects to clinical outcomes.
By constructing a unified embedding space for biomedical concepts and a wide range of common SNPs from sources such as patient-level data, biomedical knowledge graphs, and GWAS summaries, GENEREL aligns the embeddings of SNPs and clinical concepts through multi-task contrastive learning. This allows the model to adapt to diverse natural language representations of biomedical concepts while bypassing the limitations of traditional code mapping systems across different data sources.
Our experiments demonstrate GENEREL's ability to effectively capture the nuanced relationships between SNPs and clinical concepts. GENEREL also emerges to discern the degree of relatedness, potentially allowing for a more refined identification of concepts. This pioneering approach in constructing a unified embedding system for both SNPs and biomedical concepts enhances the potential for data integration and discovery in biomedical research.
\end{abstract}

\section{Introduction}

Large biobanks, such as the UK Biobank \citep{Bycroft2018TheUB}, the Million Veteran Program \citep{verma2024diversity}, and All of Us \citep{all2019all}, in conjunction with the wealth of genomic research, including genome-wide association studies (GWAS) and extensive biomedical literature, present tremendous opportunities for advancing both precision medicine and drug discovery. Insights from GWAS have already significantly enhanced our understanding of genetic predispositions to various diseases, providing critical guidance for disease diagnosis, prognosis, and treatment \citep{van2014systems}. By integrating data from large biobanks with these genomic findings, we can further expand the application of genetics in clinical settings, deepening our understanding of disease mechanisms and improving predictions of treatment responses. This synthesis of resources has the potential to revolutionize how we approach personalized healthcare.

A major challenge impeding the full potential of biobanks, genetic databases, and biomedical literature is the lack of interoperability between these resources. In particular, inconsistencies in how phenotypic traits are encoded create significant barriers to seamless data and knowledge integration. For instance, the GWAS catalog \citep{gwas} uses the Experimental Factor Ontology (EFO) \citep{efo}, while the UK Biobank maps phenotypes using SNOMED CT and Read systems. These differences make it difficult to harmonize findings across sources, as accurately mapping the same trait is often infeasible due to variations in coding systems and trait descriptions. Without addressing these interoperability challenges, the ability to fully integrate and leverage extensive genetic and phenotypic data remains limited \citep{mandl2020genomics}, hindering the broader impact of biobank and GWAS data in drug development and precision medicine.

One effective approach to harmonize diverse entities from heterogeneous data sources is to create unified representations of these concepts through representation learning, which can encode complex and heterogeneous data into a common, low-dimensional space~\citep{chen2020graph}. However, most graph-based representation learning methods rely on a large number of observed pairwise relationships between entities, which are often unavailable for phenotypic traits across sources due to differences in encoding. Even when mappings exist, they may be inaccurately aligned due to variations in hierarchy and granularity across coding systems.

A crucial yet often overlooked aspect is that all biomedical codes are accompanied by descriptions in natural language, presenting an opportunity to leverage language models for concept representation. ChatGPT \citep{openai2024gpt4technicalreport} exemplifies the remarkable natural language understanding capabilities of these models. In recent years, several language models have been developed specifically to embed biomedical concepts, trained on specialized corpora, and using various approaches. Notable examples include BioBERT \citep{Lee2019BioBERTAP}, ClinicalBERT \citep{alsentzer-etal-2019-publicly}, PubMedBERT \citep{Gu2020DomainSpecificLM}, SapBERT \citep{liu-etal-2021-self}, and CODER \citep{YUAN2022103983}. These models have shown great promise in enhancing the representation of biomedical concepts by integrating both natural language text and domain-specific knowledge, leading to more accurate and contextually informed embeddings.

While existing biomedical language models are powerful, they primarily rely on large-scale text data with limited integration of the biological mechanisms underlying clinical traits. For instance, although type 1 diabetes and type 2 diabetes both manifest with elevated blood glucose levels, the former is an autoimmune disorder characterized by decreased insulin production, while the latter involves insulin resistance typically resulting from lifestyle factors. However, embeddings for type 1 diabetes and type 2 diabetes generated by models like PubMedBERT show a cosine similarity as high as 0.995, reflecting excessive overlap despite their distinct biological mechanisms. Furthermore, these models lack the ability to represent critical genetic information, such as single-nucleotide polymorphisms (SNPs), and none of them currently provide joint representations that integrate both biomedical terms and genetic concepts at the variant level.

To address the challenges of integrating genetic and biomedical knowledge, we propose \textbf{GEN}omic \textbf{E}ncoding \textbf{RE}presentation with \textbf{L}anguage Model (GENEREL), a unified representation framework that bridges the gap between these domains. GENEREL leverages language models to encapsulate biomedical concepts based on their descriptions, generating embeddings that are collaboratively fine-tuned using diverse sources of summary-level data. These include biomedical knowledge graphs from PrimeKG \citep{Chandak2022BuildingAK} and UMLS \citep{Bodenreider2004TheUM}, patient-level data from the UK Biobank \citep{Bycroft2018TheUB}, and genomic repositories like the GWAS Catalog \citep{gwas} and Expression Quantitative Trait Loci (eQTL) \citep{Nica2013ExpressionQT}. By employing language models, GENEREL overcomes the interoperability challenges posed by heterogeneous coding systems and effectively integrates knowledge from various sources. Importantly, learning end-to-end from concept descriptions eliminates the need for anchor concepts to align information, avoiding the potential errors associated with anchor code mappings.

With the language model acting as a bridge between diverse phenotypic traits and UK Biobank, eQTL, and GWAS Catalog providing genomic information, GENEREL enriches the embeddings with a more holistic biological understanding.
To facilitate learning from multiple sources, GENEREL employs a multi-task learning paradigm. It includes three key training tasks: (1) learning relatedness from biomedical knowledge graphs in PrimeKG, (2) aligning biomedical concepts and SNPs using data from GWAS, UK Biobank, and eQTL, and (3) identifying synonyms from UMLS. 
Each of these tasks is achieved through contrastive learning~\citep{technologies9010002}, which models relatedness and aligns the embedding spaces at the same time, ensuring a cohesive representation of both genomic and biomedical concepts.
Additionally, GENEREL adjusts contrastive losses based on the relative importance of biomedical concepts and SNPs, guided by odds ratios or correlation scores. This approach ensures a comprehensive integration of information from GWAS, eQTL, and UK Biobank, making the framework highly effective for both genomic and biomedical knowledge representation.

To comprehensively evaluate the representation from our GENEREL framework, we not only employ general train-test split schemes but also extract related pairs among biomedical concepts from other biomedical databases such as DisGeNET \citep{pinero2016disgenet} and DrugBank \citep{knox2024drugbank}. 
In addition, we evaluate GENEREL SNP embedding using the GWAS results from VA's Million Veteran Program (MVP) \citep{verma2024diversity}, an independent source of genetic associations, to test the performance of GENEREL embedding across different cohort studies. 
On various benchmarks, we illustrate the state-of-the-art performance of GENEREL framework in encoding the biological relatedness between biomedical concepts and between biomedical concepts and SNPs. 
Notably, it also emerges that our embedding can also encode the different degrees of relatedness through the similarity score.
Through ablation studies, we also show that different training tasks can effectively improve the representation multifacetedly. 

The core contributions of GENEREL framework are highlighted by the following key innovative designs
that set it apart from the previous studies.
\begin{itemize}[nosep]
    \item GENEREL leverages language models to encode biomedical concepts based on their descriptions, eliminating the need for inconsistent coding systems and mappings, and enabling the representation of versatile biomedical concepts.
    \item It integrates language models and genomic variants into a unified representation framework through collaborative training to enrich the biological contexts. 
    \item The multi-task weighted contrastive learning fuses knowledge from multi-source databases, which not only grants state-of-the-art performance in detecting related concepts but also brings fine-grained relatedness levels into the representation.
    \item The unified biomedical and genomic representation system facilitates integrative biomedical research such as genetic associations, drug discovery, and personalized medicine.
\end{itemize}

\section{Related Works}
Understanding the complex connections and interactions between biomedical concepts and genomic features has long been a central focus of biomedical research. The successful outcomes of GWAS over the past few decades~\citep{gwas} have provided valuable insights into the biological underpinnings of diseases, supported clinical decision-making, and facilitated drug discovery. However, GWAS typically concentrates on a specific trait of interest and lacks the ability to generalize to multiple or broader biomedical concepts.

\noindent\textbf{Representation Learning of Biomedical and Genomic Concepts}
Existing research learns embedding for biomedical and genetic concepts by using statistical learning algorithms such as factorization of co-occurrence and adjacency matrix \citep{arora-etal-2016-latent,zhou2022multiview,hong2021clinical,gan2023arch}, or random walk based graph learning  \citep{choi2016multi,choi2016learning,zitnik2017predicting}.
Recently, researchers also applied various graph neural networks along with link prediction or graph alignment objectives to generate embeddings \citep{li2022graph}. 
Despite variations in algorithms, all the aforementioned methods are based on the codified concepts from different coding systems such as ICD10 \citep{world2004icd}, CUI \citep{Bodenreider2004TheUM}, and HPO \citep{robinson2008human}. 
Different coding systems hinder the ability of the methods to generalize across different databases. 
Manually curated code mappings between systems are needed to enable multi-source learning, which is prone to human errors~\citep{gan2023arch}. 

\noindent\textbf{Biomedical Language Models}
A wide range of pre-trained language models are employed to analyze biomedical and clinical language.
These models are trained on various domain-specific corpora such as PubMed articles \citep{yuan-etal-2022-biobart}, clinical notes \citep{huang2020clinicalbertmodelingclinicalnotes}, and knowledge graphs \citep{YUAN2022103983}. 
Masked language modeling \citep{Lee2019BioBERTAP}, next token prediction \citep{luo2022biogpt}, and contrastive learning \citep{liu-etal-2021-self} are the common techniques for adapting general language model to the biomedical domain. 
These language models have been shown to offer a more flexible and efficient method for processing biomedical knowledge \citep{10.1145/3611651}.

\section{GENEREL}

In this section, we first explain how GENEREL formalizes the task and models biomedical concepts and SNPs. We then detail the multi-task contrastive learning objective. Finally, we conclude by outlining the steps taken to extract the necessary training data from heterogeneous sources.

\subsection{Modeling}

Biomedical concepts are denoted by $\{c_i\}_{i=1}^N$, where each $c_i$ is presented by a short text phrase or description, and genomic variant concepts are denoted by $\{g_j\}_{j=1}^M$, where each $g_j$ is an indexed SNP along with the corresponding variant allele (e.g., rs2476601\_A).
We use a pre-trained language model denoted by $\mathcal{M}_\phi$ to map the biomedical concept to the dense embedding $c_i^e\in\mathbb{R}^d$, 
\begin{equation*}
    c_i^e = W_p\mathcal{M}_\phi(c_i)+b_p,
\end{equation*}
where $\phi$ represents the trainable parameters in the language model, $W$ and $b$ compose the trainable linear layer to map the hidden state from the language model into any pre-defined dimension sizes of the shared embedding space, and the hidden state is extracted from 
the [CLS] position for each concept. 
For $g_j$, since SNPs are independent concepts without any shared information, we use one-hot encoding for each $g_j$ and apply an embedding matrix $\mathcal{E}_\psi$ to generate dense representations $g_j^e\in\mathbb{R}^d$,
\begin{equation*}
    g_j^e = \mathcal{E}_\psi(g_j) = \mathcal{E}_{(\psi, j)},
\end{equation*}
where $\psi$ represents the trainable parameters in the embedding matrix.
Therefore, by using the language model, we can harmonize the biomedical concepts of text form and a preselected collection of SNPs into a unified embedding space.  

\subsection{Learning Objective}
We consider three distinct modeling tasks in GENEREL: (1) the relatedness between biomedical concepts, (2) the relationship between biomedical and genomic concepts, and (3) the disambiguation of synonyms for each biomedical concept. For each task, we can formalize the training data into a collection of concept pairs:
\begin{equation*}
\mathcal{S} \subseteq \{(h, t): h, t \in \{c_i\} \cup \{g_j\}\}.
\end{equation*}
Additionally, each pair often has an associated weight that indicates the degree of relatedness. For instance, the strength of the association between a SNP and a trait can be quantified by the odds ratio or the regression coefficient.
In GENEREL, we incorporate this information $w_{h,t}$ for a pair $(h, t)$ if available; otherwise, we set $w_{h,t}=1$.
Given the data pairs, we apply the contrastive loss to integrate the relatedness into our model. 
Specifically, we utilize the InfoNCE loss \citep{oord2019representationlearningcontrastivepredictive}:
\begin{align*}
\mathcal{L}_\mathcal{S}
&= \sum_{(h, t) \in \mathcal{S}} w_{h, t}\mathcal{L}_{\mathrm{InfoNCE}}(h, t)\\
&= - \sum_{(h, t) \in \mathcal{S}} w_{h, t} \log \frac{\exp(\mathrm{sim}(h, t)/\tau)}{\sum_{\tilde{h} \in \mathcal{C}} \exp(\mathrm{sim}(\tilde{h}, t)/\tau)}
\end{align*}
where $\mathrm{sim}(\cdot, \cdot)$ is a similarity function, $\mathcal{C}$ is the set of conditional negative samples, and $\tau$ is the temperature parameter.
In particular, we use the inner product of the embeddings as the similarity measure, i.e., $\mathrm{sim}(h, t) = \langle h^e, t^e \rangle$.
The implementation details of the InfoNCE loss may vary depending on the negative sampling schemes used \citep{oord2019representationlearningcontrastivepredictive,clip,zhang2022contrastive}. 
In GENEREL, we follow the implementation of CLIP \citep{clip} with a learnable temperature parameter. 

\subsection{Training Data}

\noindent\textbf{SNP collection}
We curate the common SNP collection from GWAS catalog and eQTL from GTEx \citep{lonsdale2013genotype}. 
Both sources compile various traits and SNP associations from existing research. 
Since our framework prioritizes common SNPs, we exclude those SNPs associated with fewer than two traits in the GWAS catalog. 
For eQTL, we retain SNPs linked to the most prevalent tissues and with the largest proportions of variance explained (PVE) values. 
We also only include SNPs with A, T, C, or G as risk alleles, omitting those with complex variants like insertions or deletions. 
Finally, we merge the selected SNPs from the GWAS catalog and eQTL, resulting in 65,278 unique SNPs and 83,900 unique genomic concepts of SNPs paired with alleles.

\noindent\textbf{GWAS catalog and eQTL}
We pair selected SNPs with their associated traits and gene names provided by GWAS catalog or eQTL. 
To enrich the data, we use the trait phrases from both the description of the original studies
and the mapped trait names from EFO coding system in GWAS catalog. 
We extract the betas or odds ratios from both sources to serve as $w_{h,t}$,
reflecting the association levels of the pairs.
These beta or odds ratio values are inconsistent across studies due to varying units,
which can differ by several orders of magnitude,
leading to unstable training.
To address this issue, we first group the pairs along with their values by both study and trait,
and then normalize the values by dividing by the mean and truncating them at specified thresholds.
All weights in the final dataset are in between $0$ and $2$. 

\noindent\textbf{UK Biobank} 
UK Biobank is a large-scale biomedical database containing participants' whole genome sequencing
together with information concerning various aspects of health. 
It is challenging to incorporate biobank patient-level data into the contrastive learning process
since the phenotype-genotype association is typically weak compared to cohort studies.
To address this issue, we first utilize the correlation matrix between phenotypes and SNPs,
adjusted for demographics such as gender and ethnicity.
We then filter the pairs by applying a threshold to the absolute correlation values.
This process can effectively identify the significantly associated phenotypes and SNPs from the patient-level data to construct high-quality training pairs. 
The correlation of each pair is also incorporated as the weight. 
We extract the EMIS cohort, consisting of 216,215 patients with 6,358 phenotypes and 61,455 SNPs.
The phenotypes are defined using PheCodes, which group ICD codes into higher-level concepts, and SNOMED CT, which are mapped to UMLS Concept Unique Identifiers (CUIs).
After processing, we have 467,026 pairs of associated concepts.

\noindent\textbf{PrimeKG} 
To further enhance the relationships among biomedical concepts, we leverage PrimeKG, a biomedical knowledge graph that contains a comprehensive array of pairwise relationships across various entities, including diseases, drugs, genes, and phenotypes.
During the training process, GENEREL primarily focuses on the biological knowledge related to diseases in PrimeKG to improve its understanding and representation of these concepts.
We filter the relationships and concepts, leaving out the rare relational types and keeping only the concept types of \textit{gene/protein}, \textit{disease}, \textit{drug}, \textit{effect/phenotype}, \textit{molecular function} and \textit{pathway}.

\noindent\textbf{UMLS}
UMLS is an integrated biomedical terminology system that serves as a useful resource for developing language models in biomedical information extraction \citep{liu-etal-2021-self,yuan-etal-2022-generative}.
UMLS concepts are organized as CUIs, which group synonymous terms that represent the same clinical concept.
In our training, we use the 2020AB release of UMLS and extract synonymous term pairs as positive samples.

Summaries and examples of the aforementioned datasets are presented in Table \ref{app:tab:example}.
\begin{table}[ht]
\centering
\caption{Summaries and examples of training data in GENEREL.}
\label{app:tab:example}
\begin{tabular}{c|cccccc}
\hline
\textbf{Task} & \textbf{Dataset} & \textbf{Example} & \textbf{Pair Number} \\
\hline
\multirow{2}{*}{Term-SNP} & GWAS\&eQTL& (colorectal cancer, rs6983267\_G, 1.180) & 135,749\\
&UK Biobank&(rheumatoid noduloses, rs1391371\_T, 0.986) & 467,026 \\
\hline
Term-Term & PrimeKG & (argatroban, cyp3a4) & 325,324\\
\hline
Synonym & UMLS & (arthritis arthritics, reiter's syndrome) & 245,812\\
\hline 
\multicolumn{3}{r}{sum.} & 1,173,911 \\
\hline
\end{tabular}
\vspace*{-10pt}
\end{table}

\section{Experiments}

\subsection{Training Setting}

For multi-task training, we utilize the processed datasets as described in the previous section.
Our language model utilizes the same architecture and initializes its weights from SapBERT \citep{liu-etal-2021-self}.
The weights and biases of the linear transformation and the SNP embeddings are initialized randomly.
For both biomedical concept and SNP embeddings, we set the dimensions to 768.
We collaboratively train GENEREL on all datasets for 25 epochs with a batch size of 512 using the AdamW optimizer with a learning rate of 2e-5 for the language model and 2e-3 for the SNP embedding matrix.
The training is performed on a single L40S GPU requiring around 40 GB of VRAM.

\subsection{Evaluation}

Our main evaluation focuses on two research questions:
\begin{enumerate}[nosep]
\item[RQ1] Can the language model effectively encoder the biological information of the biomedical concepts?
\item[RQ2] Do the language model and SNP embedding matrix form a unified representation space?
\end{enumerate}
\smallskip
\noindent\textbf{RQ1} 
For this question, we utilize associated pairs from two external biomedical knowledge bases, DisGeNET and DrugBank. DisGeNET is a platform that aggregates data on disease-associated genes and pathways from various databases and literature. We extract and sub-sample Disease-Gene and Pathway-Gene pairs from DisGeNET for evaluation. DrugBank, a key resource for pharmaceutical research, provides associations between drugs, indications, and genes. We evaluate GENEREL using Indication-Drug and Indication-Gene pairs from DrugBank.
We have confirmed that the test pairs in DisGeNET and DrugBank do not overlap with the training data from PrimeKG through exact string matching.

We compare GENEREL to several biomedical language models, including BioBERT, ClinicalBERT, PubMedBERT, SapBERT, CODER, and KRISSBERT \citep{zhang-etal-2022-knowledge}. We establish a baseline using a strong general embedding model, BGE \citep{bge_m3}. The area under the ROC curve (AUC) is evaluated for detecting related concept pairs by comparing them against randomly selected negative pairs. The similarity between concepts is measured using cosine similarity between their embeddings. 

\begin{table}[ht]
\centering
\caption{AUCs for detecting the related biomedical concept pairs against randomly sampled negative pairs. The associations include disease-gene and pathway-gene pairs from DisGeNET and Indication-Drug and Indication-Gene pairs from DrugBank. The results are reported based on 5 independent runs.}
\label{tab:lm_eval_embed}
\begin{tabular}{l|cccccc}
\hline
\multirow{2}{*}{Model} & \multicolumn{2}{c}{DisGeNET} & \multicolumn{2}{c}{DrugBank} \\
 &  Disease-Gene & Pathway-Gene & Indication-Drug & Indication-Gene \\
 \hline
 BioBERT & $0.519\pm0.013$ & $0.568\pm0.008$ & $0.714\pm0.010$ & $0.579\pm0.009$ \\
 ClinicalBERT  & $0.483\pm0.033$ & $0.528\pm0.011$ & $0.636\pm0.010$ & $0.549\pm0.009$ \\
 PubmedBERT & $0.528\pm0.023$ & $0.555\pm0.011$ & $0.711\pm0.011$ & $0.578\pm0.011$ \\
 SapBERT & $0.627\pm0.019$ & $0.585\pm0.011$ & $0.667\pm0.008$ & $0.656\pm0.006$ \\
 CODER & $0.564\pm0.015$ & $0.594\pm0.013$ & $0.811\pm0.006$ & $0.657\pm0.006$ \\
 KRISSBERT & $0.623\pm0.009$ & $0.621\pm0.010$ & $0.753\pm0.005$ & $0.745\pm0.012$ \\
 BGE & $0.640\pm0.023$ & $0.577\pm0.014$ & $0.763\pm0.005$ & $0.537\pm0.015$ \\
 \textbf{GENEREL} & $\mathbf{0.760}\pm0.023$ & $\mathbf{0.758}\pm0.009$ & $\mathbf{0.824}\pm0.009$ & $\mathbf{0.850}\pm0.005$ \\
 \hline 
 \#Pairs & 1,366 & 778 & 4,207 & 6,148 \\
 \hline 
\end{tabular}
\end{table}
As shown in Table \ref{tab:lm_eval_embed}, GENEREL achieves the highest AUCs across all four benchmarks, significantly outperforming existing baselines. This demonstrates that the GENEREL framework enables the language model to effectively encode the biological relatedness between concepts rather than purely based on their phrases or descriptions.

\noindent\textbf{RQ2}
To address this question, we evaluate GENEREL on the task of detecting associations between biomedical concepts and SNPs. We employ a standard train-test split method, using the test set from the GWAS catalog as a benchmark. Additionally, we evaluate against a genome-wide phenome-wide association study based on the Million Veteran Program (MVP) cohort~\citep{verma2024diversity}. For both benchmarks, we assess performance by calculating the AUC for distinguishing related pairs from randomly sampled negative pairs.  We define related pairs as those test split 6,718 traits and SNPs pairs for GWAS catalog and 18,141 associated traits and SNPs pairs for MVP. 

We first compare GENEREL to several conventional graph learning techniques, including TransE \citep{bordes2013translating}, TransH \citep{wang2014knowledge}, TransR \citep{lin2015learning}, DistMult \citep{yang2014embedding}, and SimplE \citep{kazemi2018simple}.
The experiments utilized an implementation of these models provided by OpenKE~\citep{han-etal-2018-openke}. For these baseline methods, we map the reported traits from the GWAS catalog to EFO codes and train them collaboratively with our curated UK Biobank pairs. To ensure consistency, we set the embedding dimension to 768 for all models, the same as in GENEREL. Both MVP and UK Biobank use PheCode to codify the trait concepts. For the MVP benchmark, we also include the matrix factorization method on our UK Biobank cohort to establish a strong baseline. We apply singular value decomposition (SVD) on the correlation matrix and keep the first 768 dimensions with the largest singular values.

\begin{table}[ht]
\centering
\caption{AUCs for detecting the related biomedical concepts and SNPs pairs on MVP and the GWAS test split. -Trait and -SNP indicates the anchors when randomly sampling negatives. Results are reported based on 5 independent runs.}
\label{tab:cross_main}
\begin{tabular}{l|cccccc}
\hline
    &  MVP-Trait & MVP-SNP & GWAS-Trait & GWAS-SNP \\
    \hline
    Cor.Mat.SVD & $0.775\pm0.009$ & $\mathbf{0.840}\pm0.004$ & - & - \\
    TransE & $0.543\pm0.015$ & $0.524\pm0.008$ & $0.693\pm0.007$ & $0.621\pm0.003$ \\
    TransH & $0.531\pm0.015$ & $0.516\pm0.004$ &$0.655\pm0.009$ &$0.601\pm0.003$\\
    TransR & $0.578\pm0.014$ & $0.528\pm0.014$ & $0.767\pm0.008$ & $0.737\pm0.008$ \\
    DistMult & $0.622\pm0.009$ & $0.761\pm0.001$ & $0.825\pm0.008$ &$0.893\pm0.002$ \\
    SimplE & $0.636\pm0.006$ & $0.759\pm0.004$ & $0.813\pm0.004$ & $0.894\pm0.001$\\
    \hline
    \textbf{GENEREL} & $\mathbf{0.793}\pm0.009$ & $0.786\pm0.004$ & $\mathbf{0.942}\pm0.003$ & $\mathbf{0.939}\pm0.002$ \\
    \hline 
\end{tabular}
\end{table}

From Table \ref{tab:cross_main},
we can see that GENEREL surpasses all the baselines by a large margin on GWAS catalog. 
On MVP, GENEREL outperforms all graph learning baselines and achieves a 0.018 improvement over SVD when using traits as anchors for random negative pairs. However, it falls short of SVD by 0.054 when negative pairs are sampled using SNPs as anchors.

Moreover, a significant drawback of all these baseline methods is their reliance on codified biomedical concepts for implementation. As mentioned earlier, different databases often employ diverse coding systems, leading to discrepancies between them. The same biomedical concepts may have different synonyms; for example, \textit{reactive arthritis} is sometimes called \textit{Reiter's syndrome}. Relying solely on codified concepts restricts models from integrating information from diverse sources and generalizing across different systems. One possible reason the graph learning baseline methods underperform compared to GENEREL is their inability to learn and harmonize information across datasets, as biomedical terms are represented by PheCode in MVP and the UK Biobank, while they are codified by EFO in the GWAS catalog. Simply merging these datasets allows message passing only through SNP concepts, which is too sparse. A key advantage of our GENEREL framework is its use of language models to encode biomedical concepts directly from language terms, thereby breaking down the barriers of codified data.

\subsection{Ablation Study}
A key feature of the GENEREL framework is its incorporation of multi-task and multi-source training.
To demonstrate the function of each training task, we conduct ablation experiments on different combinations of the training datasets.
Besides the previous benchmarks, we also include COMETA \citep{basaldella-etal-2020-cometa},
a dataset curated from public anonymous health discussions on Reddit,
to evaluate the model performance on disambiguating synonyms in biomedical concepts.  
COMETA contains 20k English biomedical mentions in various forms of daily languages. 
We pool the samples in the ``general'' and ``specified'' splits.
We report the AUCs for synonym pairs and randomly sampled negative pairs to maintain consistency with other benchmarks.
The results are listed in Table \ref{tab:ablation}.

\begin{table}[ht]
\centering
\caption{AUCs for the ablation studies on various benchmarks. - means removing the dataset from the GENEREL framework. Results are reported based on 5 independent runs.}
\label{tab:ablation}
\resizebox{\textwidth}{!}{  
\begin{tabular}{l|cc:cc:c|c}
\hline
    \multirow{2}{*}{Model Ablation} & \multicolumn{2}{c:}{Term-Term} & \multicolumn{2}{c:}{Term-SNP} & Synonym &  \multirow{2}{*}{average} \\
    & DisGeNET & DrugBank &  MVP & GWAS & COMETA &  \\
    \hline
    \textbf{GENEREL} & $0.764_{\pm0.016}$ & $0.837_{\pm0.004}$ & $0.792_{\pm0.006}$ & $0.941_{\pm0.002}$ & $0.977_{\pm0.001}$ & \textbf{0.862} \\
    -UMLS & $0.771_{\pm0.013}$ & $0.838_{\pm0.006}$ & $0.807_{\pm0.003}$ & $0.940_{\pm0.001}$ & $0.932_{\pm0.001}$ & 0.857 \\
    -UMLS-PrimeKG  & $0.683_{\pm0.022}$ & $0.737_{\pm0.005}$ & $0.815_{\pm0.006}$ & $0.950_{\pm0.002}$ & $0.944_{\pm0.002}$ & 0.826 \\
    -UMLS-PrimeKG-GWAS & $0.670_{\pm0.014}$ & $0.690_{\pm0.007}$ & $0.620_{\pm0.008}$ & $0.549_{\pm0.007}$ & $0.922_{\pm0.001}$ & 0.690 \\
    \hline 
\end{tabular}
}
\end{table}

Without the UMLS training task, we observe a decline in performance on the COMETA benchmark, as the model's ability to disambiguate synonyms decreases due to the lack of synonym information in the other datasets.
When further excluding PrimeKG from training, the performance on DisGeNET and DrugBank drops by 0.098 and 0.101 respectively. 
The GWAS catalog and UK Biobank primarily focus on gene and trait concepts, lacking broader biomedical concepts such as pathways and drugs.
PrimeKG enhances the model's learning by integrating this additional information.
When trained only on UK Biobank, the model performs worse uniformly across the benchmarks,
since GWAS covers a broader range of biomedical concepts and SNPs compared to UK Biobank.
Overall, the ablation study demonstrates the necessity and functionality of each training task,
showing the benefits of the multi-task, multi-source training scheme.

\section{Discussion}

\subsection{Encoding the Relative Relatedness Levels}

When modeling the relatedness between biomedical concepts and SNPs, an important fact is that the degrees of how SNPs influence concepts such as traits are different. 
In general, GWAS reflects this through the difference in the odds ratio or beta coefficients from regression. 
We consider and infuse this information into GENEREL through our weighted InfoNCE loss. 

Here we discuss to what extent the representation from GENEREL can encode the relative relatedness. 
Given an anchor trait, we pair two associated SNPs with relative differences in odds ratios to form a test sample. 
We also group these test samples by the difference in odds ratios in a monotonic manner.
We conduct evaluation on the GWAS test split and the MVP database mentioned before. 

\begin{figure}[t]
\centering
\includegraphics[width=0.5\textwidth]{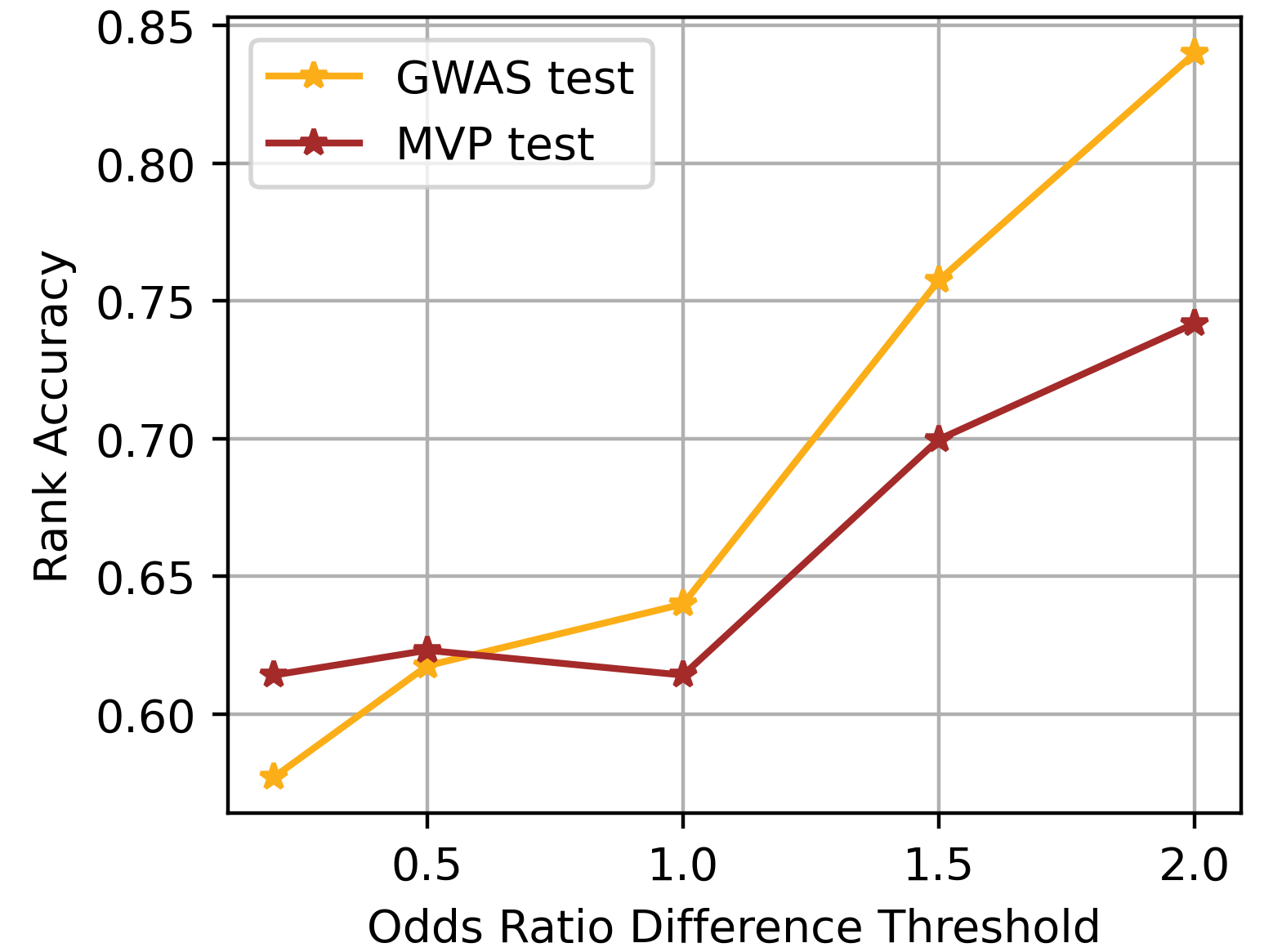}
\caption{Evaluation of detecting the relative degree of relatedness.
The plot depicts the accuracies of different sub-sample groups with various odds ratio differences.}
\label{fig:rank}
\vspace{-10pt}
\end{figure}

As illustrated in Figure~\ref{fig:rank}, GENEREL embedding effectively captures the relative levels of relatedness; as the differences become more pronounced, the performance of our embedding improves. On the GWAS test, on the samples with odds ratio gaps larger than 2, GENEREL can achieve nearly 85\% accuracy.
These results demonstrate that GENEREL can more accurately detect samples with greater differences in association degrees (larger differences in odds ratios).
Hence, GENEREL shows the ability to encode the degree of association between traits and SNPs, validating the effectiveness of the weighted contrastive loss.

\subsection{Robustness to Synonyms}
To test how robust our GENEREL framework is against synonyms in biomedical concepts, we construct two sub-sampled test sets from MVP and the GWAS catalog test split. We use the synonyms from previous research \citep{wu2019mapping,mcarthur2023linking} for MVP and from UMLS for GWAS to substitute the original term phrases with synonyms. 

\begin{table}[ht]
\centering
\caption{AUCs of GENEREL detecting the concept-SNP relatedness against random negatives on the original concept phrases and the substituted synonyms on sub-sampled MVP and the GWAS catalog test split.}
\label{tab:syn}
\begin{tabular}{c|cc}
\hline
 & MVP & GWAS \\
 \hline
   Original  & $0.798_{\pm0.008}$ & $0.901_{\pm0.004}$  \\
   Synonyms  & $0.786_{\pm0.005}$ & $0.836_{\pm0.005}$  \\
\hline
\end{tabular}
\end{table}
As shown in Table \ref{tab:syn}, although the performance fluctuates when evaluating synonyms, GENEREL can still effectively detect the biomedical concept and SNP associations. 
This verifies that GENEREL performs robustly against the changes of synonyms in biomedical concepts.

\subsection{Case Study}
To further demonstrate the performance of our model, we visualize and compare the embeddings generated by GENEREL and PubMedBERT. Using t-SNE \citep{van2008visualizing} to reduce the dimensionality to two, we create visual representations of the embeddings. Additionally, for GENEREL, we include embeddings for SNPs.

\begin{figure}[t]
\centering
\includegraphics[width=\textwidth]{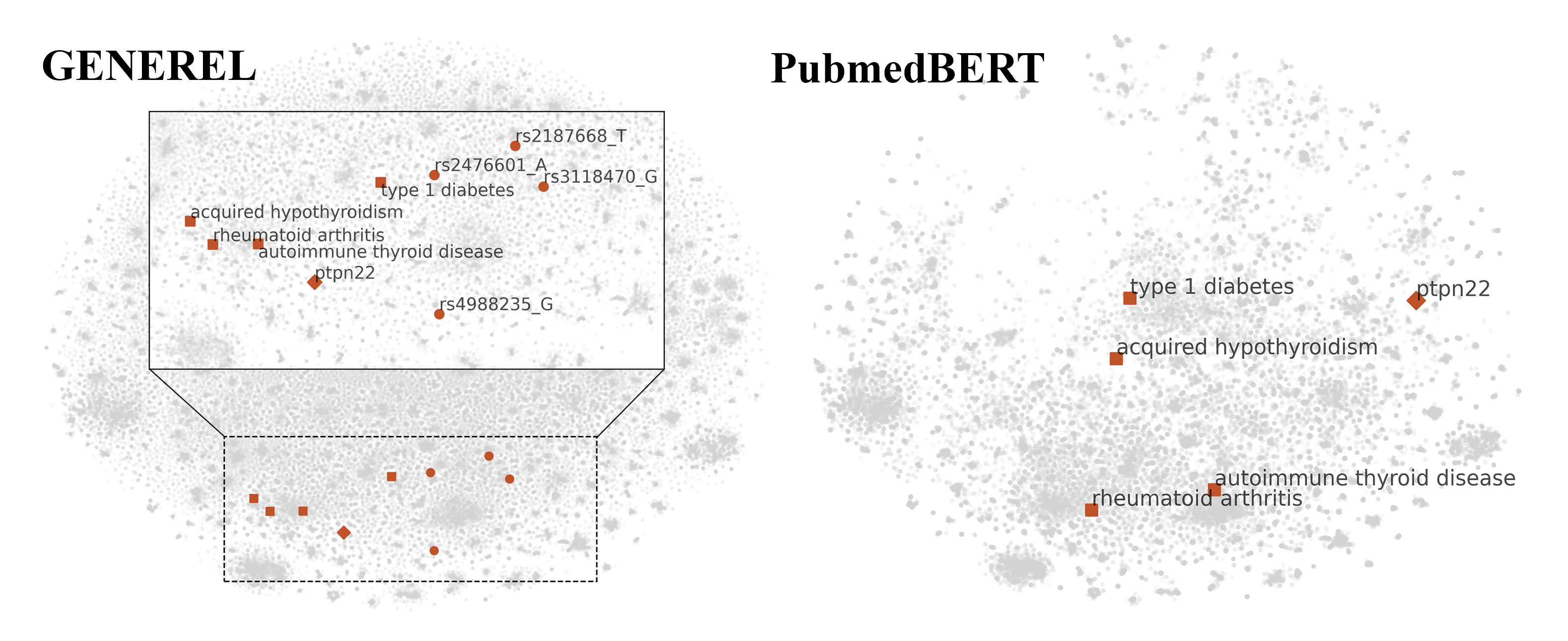}
\caption{Embedding visualizations of GENEREL and PubmedBERT using t-SNE. We highlight the autoimmune diseases and their associated genes, as well as the relevant SNPs for GENEREL.}
\label{fig:case}
\vspace{-10pt}
\end{figure}

Autoimmune diseases like type 1 diabetes, autoimmune thyroid disease, and rheumatoid arthritis affect a large portion of the population, making them a major public health concern and a frequent focus of research worldwide \citep{parameswaran2015identification}. These diseases have complex mechanisms, and studies have confirmed that type 1 diabetes and rheumatoid arthritis are linked to mutations in the PTPN22 gene \citep{bottini2006role}. In Figure \ref{fig:case}, the left plot shows that GENEREL effectively clusters the autoimmune diseases and related gene concepts into a localized group, whereas the embeddings from PubMedBERT (right plot) are more dispersed across the space. Additionally, for GENEREL, we highlight relevant SNPs and risk alleles connected to these biomedical concepts. For example, rs2476601\_A is from PTPN22, while rs2187668\_T, rs3118470\_G, and rs4988235\_G are all associated with these diseases. As shown in the left plot, these SNPs and risk alleles are tightly grouped within the GENEREL embeddings, demonstrating the model's ability to capture biological relatedness.

In the other case study, we focus on comparing two semantically similar diseases. As we mentioned before, type 1 diabetes shares similar symptoms as type 2 diabetes, however, the pathogenic mechanisms of the two diseases are different. 
While type 1 diabetes is an autoimmune disease which is usually caused by genetics and exposure to viruses and other environmental factors, the risk factors for type 2 diabetes include obesity, age, and family history.
The genetic factors of the two diseases are quite distinct.
In PubmedBERT embeddings
the cosine similarity between the two concepts is excessively high at 0.995,
indicating a high relatedness, while in GENEREL it is adjusted to a lower 0.815,
showing the model's ability to capture more biological information. 

\section{Conclusion}

In this paper, we proposed GENEREL, a framework that incorporates language models to encode the biomedical concepts from their phrases or descriptions, collaboratively with a broad set of common SNPs. 
This design alleviates the framework's reliance on various coding systems to represent concepts and bypasses the limitations of traditional code mappings, facilitating learning across diverse data sources.
To that end, GENEREL is empowered with multi-task, multi-source contrastive learning tasks, infusing information from biomedical knowledge graphs, GWAS catalog, and patient-level data of different institutions.
Through extensive evaluations, we quantitatively and qualitatively demonstrate state-of-the-art performance in modeling the association between biomedical concepts and genomic variants and the capability of learning across data sources. 
GENEREL also shows to discern the degree of relatedness between concepts, allowing a more nuanced identification of associations. 
Overall, GENEREL presents a pioneering framework in joint representation learning of genomic and biomedical concepts. 
It can facilitate and enhance the integration, discovery, and understanding of the biological mechanism in biomedical research.

\end{document}